\documentclass[letterpaper, 10 pt, conference]{ieeeconf}  

\IEEEoverridecommandlockouts                              

\usepackage{hyperref}
\usepackage{graphics} 
\usepackage{epsfig} 
\usepackage{mathptmx} 
\usepackage{times} 
\usepackage{amsmath} 
\usepackage{amssymb}  
\usepackage[linesnumbered,ruled,vlined]{algorithm2e}
\usepackage{algorithmic}
\RestyleAlgo{ruled}
\usepackage{tabularx,booktabs}

\usepackage[dvipsnames]{xcolor}
\definecolor{new}{HTML}{000000}
\definecolor{update}{HTML}{000000}

\title{\LARGE \bf
Gaussian Splatting Visual MPC for Granular Media Manipulation
}

\author{Wei-Cheng Tseng$^{1}$$^{2}$, Ellina Zhang$^{1}$,  Krishna Murthy Jatavallabhula$^{3}$ and Florian Shkurti$^{1}$$^{2}$
\thanks{Corresponding author email: \texttt{weicheng.tseng@mail.utoronto.ca }}
\thanks{$^{1}$ University of Toronto, $^{2}$ Vector Institute, $^{3}$ MIT CSAIL}%
}

\begin{document}

\maketitle
\thispagestyle{empty}
\pagestyle{empty}

\newcommand{\cvec}[1]{\textbf{\textit{#1}}}
\begin{abstract}
%
%
Recent advancements in learned 3D representations have enabled significant progress in solving complex robotic manipulation tasks, particularly for rigid-body objects. However, manipulating granular materials such as beans, nuts, and rice remains challenging due to the intricate physics of particle interactions, high-dimensional and partially observable state, inability to visually track individual particles in a pile, and the computational demands of accurate dynamics prediction. Current deep latent dynamics models often struggle to generalize in granular material manipulation due to a lack of inductive biases. In this work, we propose a novel approach that learns a visual dynamics model over Gaussian splatting representations of scenes and leverages this model for manipulating granular media via Model-Predictive Control. Our method enables efficient optimization for complex manipulation tasks on piles of granular media. We evaluate our approach in both simulated and real-world settings, demonstrating its ability to solve unseen planning tasks and generalize to new environments in a zero-shot transfer. We also show significant prediction and manipulation performance improvements compared to existing granular media manipulation methods.
\end{abstract}


\section{Introduction}
Neural rendering and view synthesis methods~\cite{eccv20_nerf,neurips19_srn,cvpr19_occnet,23_gaussian_splatting,cvpr19_deepsdf,icra22_clanerf} have enabled a wide set of applications in scene understanding, 3D reconstruction, and representation. Moreover, they have shown promise in many complex robot manipulation tasks on rigid body objects~\cite{graspnerf,rss21_giga}. Manipulating granular materials such as beans, nuts, rice, oats, and other such objects common in daily life remains a challenging problem, so in this paper, we address the question of whether neural rendering methods, Gaussian Splatting in particular, provide a good representation for control of granular media. 

\begin{figure}[t]
    \centering
    \includegraphics[width=1\linewidth]{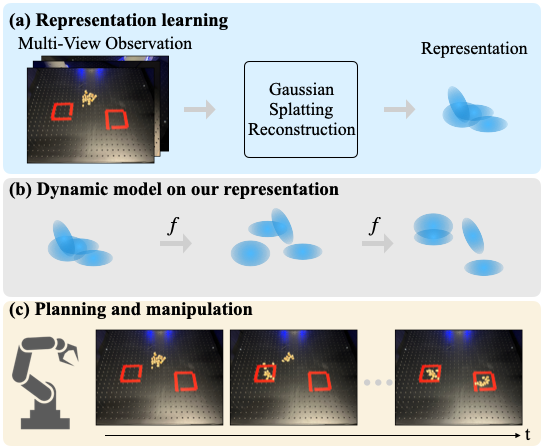}
    \caption{Our method takes a few multi-view images of a scene and their corresponding camera poses as input, and (a) converts them into their Gaussian splatting representation, (b) learns a dynamics model $f$ over these representations, and (c) performs visual model-predictive control for granular material manipulation, which requires view synthesis and dynamics rollouts.}
    \label{fig:firstfig}
\end{figure}

Several factors contribute to the difficulty of granular material manipulation. First, modeling the interactions between particles is complicated due to the intricate physics involved~\cite{iclr19_particle_dynamics} and the unknown geometry of individual particles. Second, accounting for all particles in planning requires a high-dimensional state~\cite{corl17_granular,ral21_gan_mani}, which creates challenges for downstream policy learning or planning algorithms.
Third, visually identifying and tracking individual granular particles in a pile is nearly impossible due to their self-similarity, which leads to data association problems, and due to the inherent partial observability of the setting. This has led to the majority of existing works that simulate granular materials without ground truth data from the real world~\cite{iclr19_dpi,iclr24_3dparticle,neurips22_plasticnet}, relying on simulated data. Finally, accurately predicting particle dynamics is computationally expensive~\cite{iclr19_particle_dynamics,siggraph16_mpm,rss23_dyna_resolution}.

To address these challenges, recent efforts have focused on modeling the visual dynamics of granular piles using highly expressive neural network latent dynamics models directly from pixels~\cite{wafr20_}. However, these models often underperform compared to linear dynamics models due to a lack of inductive biases. In contrast, physics-inspired approaches, such as particle-based models, introduce strong inductive biases for neural network dynamics models. 

In this work, we show that Gaussian splatting~\cite{23_gaussian_splatting} over video frames provides effective representations for downstream model-predictive control over granular media. 
Gaussian splatting is an image rendering and reconstruction technique originating in computer graphics. It represents a 3D scene as a collection of Gaussians (splats), each centered around points in the scene and associated with a single color. From a given viewpoint, an image of the scene is rendered by projecting each splat into 2D and blending colors where they overlap. 
\textcolor{new}{
This collection of splats that can also be parameterized by the position and rotation of Gaussians provides a smooth and continuous representation of the scene, making it particularly well-suited for modeling video sequences.
}

\textbf{Our contribution:} We use the Gaussian splats representing the scene at each time as a state vector that can be manipulated via MPC, effectively lowering the dimensionality of the image. We learn a dynamics model over Gaussian splats and show that by doing MPC with this dynamics model and representation, robots can efficiently handle complex and precise manipulation tasks involving granular materials. This representation enables robots to optimize their actions, anticipate challenges, and adapt to dynamic environments.

\textcolor{update}{
We evaluate our approach across a range of granular material manipulation tasks in both simulation and real-world settings, demonstrating its superiority over existing baselines in dynamics prediction and task performance. Our model successfully enables solutions of complex planning tasks. Furthermore, we highlight its generalization capability by transferring a trained model to new environments with varying object shapes in a zero-shot setting.
}
\footnote{For more details, code, videos, and the paper's appendix, please refer to our project website: \href{https://rvl.cs.toronto.edu/gs-granular-media-mpc/}{https://rvl.cs.toronto.edu/gs-granular-media-mpc/}.}

\section{Related Work}
\label{sec:citations}

\textbf{3D Visual Representations for Manipulation.} The use of 3D visual representations~\cite{eccv20_nerf,neurips19_srn,cvpr19_occnet,23_gaussian_splatting,cvpr19_deepsdf,icra22_clanerf} for robotic manipulation has gained significant traction in recent years. One of the foundational approaches in this domain involves the use of 3D point clouds, which provide a detailed geometric representation of objects in the environment. Works such as PointNet~\cite{cvpr16_pointnet} and PointNet++~\cite{cvpr17_pointnetpp} have been pivotal in processing and understanding 3D point clouds, enabling robots to perform tasks like object recognition and grasping~\cite{corl22_framemining}.
Voxel-based representations~\cite{cvpr19_occnet} have also been widely explored for manipulation tasks. By discretizing the workspace into a grid of voxels, these methods offer a straightforward way to model the occupancy and structure of the environment. 
\textcolor{update}{
VoxNet~\cite{iros15_voxnet} introduced a deep learning architecture that uses 3D voxel grids for object recognition in robotic tasks. Similarly, a voxel-based deep Q-network~\cite{cvpr22_c2f,icra22_arm} was developed for robotic grasping, which demonstrated the effectiveness of 3D voxel representations in manipulation tasks.
}

Another prominent line of research involves the use of implicit neural representations, where 3D shapes and environments are encoded as continuous functions rather than discrete points or voxels. Neural Radiance Fields (NeRF)~\cite{eccv20_nerf,icra22_clanerf} is a notable example of this approach, where scenes are represented as volumetric radiance fields that can be rendered from arbitrary viewpoints. While originally designed for view synthesis, NeRF and its variants have inspired applications in robotic manipulation, especially in scenarios where precise modeling of object geometry and appearance is critical~\cite{corl23_feature_field,icra22_nerf_supervision,yen2022mira}.



\textbf{Particle Dynamics.} 
\textcolor{update}{
%
Learning-based dynamics models commonly incorporate inductive biases, particularly object-centric representations.
Particle-based representations have been shown to have strong inductive biases when representing deformable objects. 
For instance, DPI-Net~\cite{iclr19_dpi} combines a hierarchical particle dynamics module with MPC-based control for deformable object manipulation. 
However, particle-based approaches suffer from scalability issues as the number of particles increases~\cite{icml20_gns,kumar2022gnsgeneralizablegraphneural}, which makes them computationally expensive and impractical in planning tasks. 
}

\textbf{Granular Material Manipulation.} Manipulating granular media by pushing piles of small objects into a desired target set using visual feedback has been accomplished with models as simple as linear~\cite{wafr20_,ral21_gan_mani}. Granular material manipulation presents unique challenges due to the complex and non-linear interactions of these materials~\cite{corl23_neural_field}. Traditional approaches often rely on physics-based models~\cite{iclr19_particle_dynamics}, which simulate individual particles to predict bulk material behavior. While accurate, these models are computationally intensive and may not be suitable for real-time manipulation.

To address these limitations, recent research has explored data-driven methods that learn granular dynamics from observations~\cite{cvpr23_physgau}. Techniques like neural networks have been employed to approximate material behavior~\cite{corl23_neural_field,rss23_dyna_resolution,wafr20_}, enabling faster predictions during manipulation tasks. However, these approaches often struggle with generalization across different types of granular materials and varying conditions. Besides, these granular material manipulations only represent material in 2D space, which constrains the potential of 3D manipulation tasks. Our approach leverages an advanced 3D reconstruction technique, which alleviates this limitation. 


\begin{figure*}
    \centering
    \includegraphics[width=1\linewidth]{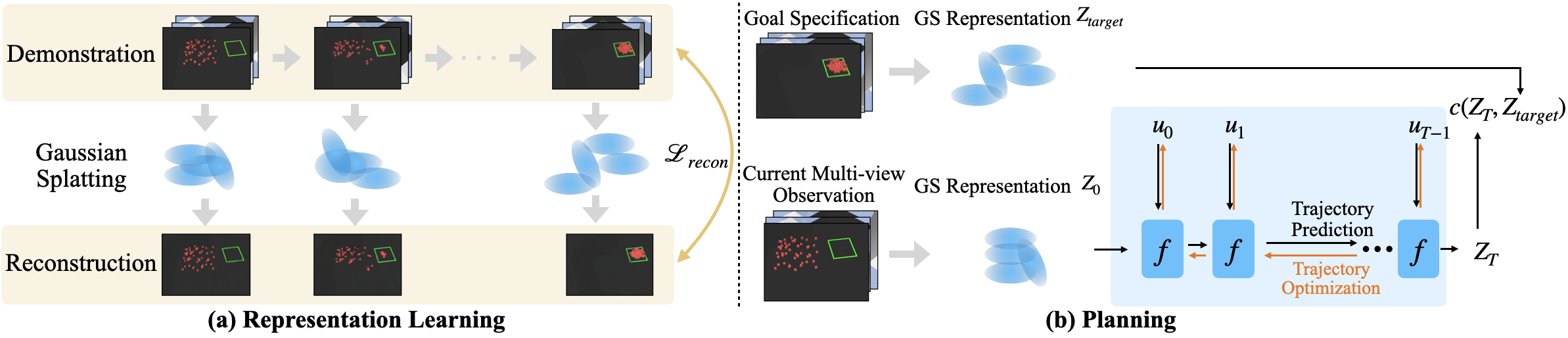}
    \caption{
    \textcolor{update}{
    \textbf{Our framework.}
    (a) Given demonstration trajectories with multi-view observations, we leverage Gaussian splatting representations to reconstruct the observed images at each timestep.
    (b) The dynamics model $f$ predicts the temporal evolution of the Gaussian Splatting representation $Z_t$ with input action ${\cvec{u}_t}$. In the planning stage, we optimize the action sequence ${\cvec{u}_t}$ by minimizing the task objective $c(\cvec{Z}_T, \cvec{Z}_{target})$.
    }
    }
    \label{fig:overview}
\end{figure*}

\section{Preliminaries}
Gaussian splatting~\cite{23_gaussian_splatting} has emerged as a powerful rendering technique that can capture the state of the visual world with a discrete set of 3D Gaussians $G = \{n^i\}$, where $n^i = (\cvec{g}^i,  \cvec{s}^i, \cvec{R}^i, \cvec{s}^i, \sigma^i, \cvec{c}^i)$ represents a 3D Gaussian.
Each Gaussian $i$ is parameterized by its position $\cvec{g}^i \in \mathbb{R}^3$, orientation $\cvec{R}^i \in \mathbf{SO}(3)$, scale $\cvec{s}^i \in \mathbb{R}^3$, opacity $\sigma^i \in \mathbb{R}^+$, and color $\cvec{c}^i \in \mathbb{R}^3$.
Given a viewpoint whose transform relative to the world frame is denoted by $\cvec{V} \in \mathbf{SE}(3)$ and projection function from the 3D world to the view’s screenspace is defined by $\pi(\cvec{x})$, the color at a pixel coordinate $\cvec{p}$ can be calculated by sorting the Gaussians in increasing order of their viewspace z-coordinate and then using the splatting formula:
\begin{equation}
    C_\text{RGB}(\cvec{p}) = \sum_{i\in N} \cvec{c}^i \alpha^i(\cvec{p}) \prod_{j=1}^{i-1} (1-\alpha^i(\cvec{p}))
\end{equation}
\begin{equation}
    \alpha^i(\cvec{p}) = \sigma_i \, \text{exp}(g_i(\cvec{p}))\;,
\end{equation}
\begin{equation}
    {g}_i(\cvec{p}) = \mathbf{x}^{iT} \hat{\Sigma}^{-1}_i \mathbf{x}^i\;, 
    \mathbf{x}^i = \cvec{p} -\pi(\cvec{g}^i)
\end{equation}
$\hat{\Sigma}_i = \mathbf{J}\mathbf{V} {\Sigma}_i \mathbf{V}^T \mathbf{J}^T$
is the covariance of the Gaussian $i$ projected into the viewpoint’s screenspace where $\mathbf{J}$ is the Jacobian of the projection function $\pi(\cvec{x})$ and $\Sigma_i = \cvec{R}_i \; \text{diag}(\cvec{s}^2_i) \; \cvec{R}^T_i$. Further details about this process can be found in~\cite{23_gaussian_splatting}. 

The parameters of the Gaussians can be updated by minimizing the $\mathcal{L}_1$ loss and structural similarity index measure (SSIM) $\mathcal{L}_{SSIM}$ between the reconstructed image and ground-truth image. 

\begin{equation}
\begin{aligned}
        \mathcal{L}_{recon}(I_{recon}, I_{GT}) = &\mathcal{L}_1(I_{recon}, I_{GT})+ \\
        & \beta (1-\mathcal{L}_{SSIM}(I_{recon}, I_{GT}))
\end{aligned}
\end{equation}
\noindent where $I_{GT}$ and $I_{recon}$ are ground-truth and reconstructed images (with reconstructed pixel $\cvec{p}$ having color $C_\text{RGB}(\cvec{p})$), respectively.
Note that SSIM is calculated using various windows of an image. The measure between two windows $x$ and $y$ of common size $N \times N$.
\begin{equation}
    SSIM(x, y) = \frac{(2\mu_x\mu_y + c_1)(2\sigma_{xy}+c_2)}{(\mu_x^2 + \mu_y^2 + c_1)(\sigma_x^2 + \sigma_y^2 + c_2)}
\end{equation} 
\noindent where $\mu_x$ and $\mu_y$ represent pixel mean of $x$ and $y$. $\sigma_x$ and $\sigma_y$ indicate pixel variance of $x$ and $y$. $\sigma_{xy}$ is the covariance of $x$ and $y$, and $c_1$ and $c_2$ are constant to stabilize division.

\section{Our Approach}

\subsection{Problem Formulation} \label{sec:pf}
Given multi-view RGBD observations $\cvec{O}_{target} = \{\cvec{o}^v, m^v\}_{v=1}^N$ of the target pattern of the granular material, where $o^v$ represents the RGBD image and  $m^v$ indicates the corresponding camera pose, we would like to manipulate the granular material to minimize the cost measurement $c$ such that the granular material can match the target pattern. 
\textcolor{update}{The trajectory optimization problem over a horizon $T$ can be defined as follows:}

\begin{equation}
    \label{equ:optm}
    \cvec{u}_{0:T-1} = argmin_{\cvec{u}_{0:T-1}} c(\cvec{Z}_T, \cvec{Z}_{target})\\\
\end{equation}
\begin{equation}
    \cvec{Z}_0 = h(\cvec{O}_0), \,\,\cvec{Z}_{target} = h(\cvec{O}_{target}), \; \cvec{Z}_{t+1} = f(\cvec{Z}_t, \cvec{u}_t)
\end{equation}
where $h$ is the perception module that performs Gaussian splatting and \textcolor{update}{$f(\cdot)$} is the dynamic model which predicts the representation’s evolution $\cvec{Z}_{t+1}$ from the previous representation and action. This optimization process aims to acquire the sequence of actions $\{\cvec{u}_t\}$ to minimize the cost function $c(\cvec{Z}_T, \cvec{Z}_{target})$. Also, $\cvec{u}_t \in \mathbb{R}^4$ represents the starting position and pushing direction of the end-effector. To enable learning a dynamics model for granular materials, we collect a dataset $D_\text{RGBD} = \{(\cvec{O}_t, \cvec{u}_t, \cvec{O}_{t+1})\}$ via a simulator of a manipulator interacting with particles.

\subsection{Representation of Granular Materials} \label{subsec:repr}
%
%
%

We leverage Gaussian splatting~\cite{23_gaussian_splatting} as the representation method for image observations of granular materials. Instead of initializing the Gaussian splats with a pointcloud formed from structure-from-motion, as is typically done in 3D scene reconstruction applications of neural rendering, 
we lift the RGBD image with the corresponding camera pose to 3D space and form the point cloud.
To ease the downstream training of dynamic models, we downsample the point clouds used by initializing the Gaussian splatting with farthest point sampling~\cite{cvpr17_pointnetpp}, which iteratively samples the farthest point and performs distance updating.
Given the multiview observation $\{\cvec{O}_t\}$, we leverage Gaussian splatting for reconstruction $\{\cvec{G}_t\}$. 
Since we are only interested in granular materials, we remove the Gaussian that represents the background or the ones that have high transparency. We then denote the remaining Gaussians as $\{\cvec{Z}_t\}$.
More specifically, we transform the dataset $D_{RGBD}$ to $D_{GS} = \{(\cvec{Z}_t, \cvec{u}_t, \cvec{Z}_{t+1})\}$ by training Gaussian splatting reconstruction for each image frame.
\subsection{Learning a Visual Dynamics Model} \label{subsec:dynamics}
Given the dataset $D_{GS}$, we learn a visual dynamics model over Gaussian splatting representations.
Our dynamics model $f(\cdot)$ is structured as a Graph Neural Network (GNN)~\cite{neurips17_sage} with iterative message passing that takes the Gaussian splatting reconstruction as input and predicts the translation and rotation for each Gaussian.

To be more specific, to transform Gaussian splitting $\cvec{Z}_t$ into a graph structure $\mathcal{G}_t = (\mathcal{V}_t, \mathcal{E}_t)$, where $\mathcal{V}_t = \{v^i_t\}_{i=1...|\mathcal{V}_t|}$ indicate a set of nodes and $\mathcal{E}_t = \{e^i_t\}$ represent a set of edges.
we create a graph by adding edges between Gaussians if the $L_2$ distance between the respective vertices is smaller than a distance threshold $\omega$. We form the node features of the GNN as $(\cvec{c}^i_t, \sigma^i_t, \cvec{R}^i_t, \cvec{g}^i_t, \cvec{s}^i_t)$ for node $\cvec{v}^i_t$. $f$ consists of node encoder $f_{enc}$ with node representation $\Bar{v}^i$ from $v^i_t$:
\begin{equation}
    \Bar{\cvec{v}}^i_t = {f_{enc}(\cvec{v}^i_t,\, \cvec{u}_t)}
\end{equation}
Then, we have a message-passing encoder $f_{msg}$ which allows us do multi-step message passing:
\begin{equation}
    \cvec{q}^{i, \gamma+1}_t = f_{msg}(\cvec{q}^{i, \gamma}_t, \text{mean}_{j \in N_i} \cvec{q}^{j, \gamma}_t),\;\; \cvec{q}^{i, 0}_t = \Bar{\cvec{v}}^{i}_t
\end{equation}
where $N_i$ is a set of nodes that connected to node $i$. Finally, we have the decoder $f_{dec}$ that transforms node features after $\Gamma$ message passing steps to dynamic information
\begin{equation}
    \Delta \cvec{r}^i_t\,, \Delta \cvec{g}^i_t = f_{dec} (\cvec{q}^{i, \Gamma}_t)
\end{equation}

\noindent where $\Delta \cvec{r}^i$ and $\Delta \cvec{g}^i$ indicates the displacement and rotation of Gaussian $\cvec{n}^i$. We then move and rotate the Gaussian:
\begin{equation}
    \hat{\cvec{g}}^i_{t+1} = \cvec{g}^i +  \Delta \cvec{g}^i_t, \;\; 
    \hat{\cvec{r}}^i_{t+1} = \Delta \cvec{r}^i \cdot \cvec{r}^i_t 
\end{equation}
\noindent where $\cvec{r}$ is the quaternion representation of $\cvec{R}$. In the end, we obtain a set of Gaussians that represents the next image:
\begin{equation}
    \hat{\cvec{Z}}_{t+1} = \{(\cvec{c}^i_t, \alpha^i_t, \hat{\cvec{R}}^i_{t+1}, \hat{\cvec{g}}^i_{t+1}, \cvec{s}^i_t) \}
\end{equation}


%
%
\noindent Then, we use the Chamfer distance to train the dynamics:
\begin{equation}
    \begin{aligned} 
    \mathcal{L}_{dyna} =& \frac{1}{|\hat{\cvec{Z}}_{t+1}|} \sum_{\hat{\cvec{n}}_{t+1} \in \hat{\cvec{Z}}_{t+1}}\min_{\cvec{n}_{t+1} \in \cvec{Z}_{t+1}}  \mathcal{L}_{gaussian}(\hat{\cvec{n}}_{t+1}, \cvec{n}_{t+1})  \\ 
    &+ \frac{1}{|\cvec{Z}_{t+1}|} \sum_{\cvec{n}_{t+1} \in \cvec{Z}_{t+1}}\min_{\hat{\cvec{n}}_{t+1} \in \hat{\cvec{Z}}_{t+1}}  \mathcal{L}_{gaussian}(\cvec{n}_{t+1}, \hat{\cvec{n}}_{t+1})
    \end{aligned}
\end{equation}
\begin{equation}
    \mathcal{L}_{gaussian}(\cvec{n}, \hat{\cvec{n}}) = ||\cvec{g} - \hat{\cvec{g}}||_2 + \lambda 
    (1 - |\cvec{r}_{t+1} \cdot \hat{\cvec{r}}|)
    %
\end{equation}
where $\lambda$ is a hyperparameter that determines the importance between position and orientation.

\subsection{Planning} \label{subsec:planning}
\noindent Inspired by \cite{iclr24_dreamgau}, we leverage the density field 
\begin{equation}
    d(\cvec{x}) = \sum_{
    \cvec{n}_i \in \cvec{Z}} \sigma_i \cdot \text{exp}( (\cvec{x}-\cvec{g}_i)^T \Sigma_i^{-1} (\cvec{x}-\cvec{g}_i))
\end{equation}
which indicates whether a specific 3D position $x$ is occupied by any material and we use it to form the cost function for the planning algorithm. The cost function used in the planning algorithm is the following:
\begin{equation}
    c(\cvec{Z}_t, \cvec{Z}_{target}) = \frac{1}{\lvert P \rvert} \sum_{\cvec{x} \in P} \lvert d_t(\mathbf{x}) - d_{target}(\cvec{x}) \rvert ^2
\end{equation}
where $P$ is a pre-defined set of points we would like to query. This cost helps us measure the difference between the occupied space in $\cvec{Z}_t$  and that in $\cvec{Z}_{target}$.

\textcolor{update}{
We perform the optimization shown in Equation \ref{equ:optm} as part of a gradient-based MPC loop (as shown in Alg.~\ref{alg:planning}) and we execute the optimized action sequence in a closed-loop fashion. 
After the first action in the sequence is executed by the robot, we re-run the planning algorithm to generate a new action sequence.
}


\SetKwComment{Comment}{/* }{ */}
\begin{algorithm}
    \caption{Our Visual MPC Planning Algorithm}\label{alg:planning}
    \KwData{Current observation $\cvec{O}_t$, target $\cvec{O}_{target}$, planning horizon $T$, 
    the dynamics model $f$, Number of sampled action sequence $K$ and gradient descent step $N$}
    \KwResult{a sequence of actions $\cvec{u}_{0:T-1}$}
    \

    Get current representation $\cvec{Z}_t$ from observation $\cvec{O}_t$\;
    Acquire target representation $\cvec{Z}_{target}$ from $\cvec{O}_{target}$\;

    Sample $K$ initial action sequences $\cvec{u}^{1:K}_{0:T- 1}$ \;
    \

    \For{k $\leftarrow$ 1 \text{to} K}{
        \For{i $\leftarrow$ 1 \text{to} N}{
            \For{t $\leftarrow$ 0 \text{to} T-1}{
                Predict next representation $\cvec{Z}_{t+1} = f(\cvec{Z}_{t}, \cvec{u}_t)$\;
            }
            Compute the task loss $c^k = c(\cvec{Z}_{T}, \cvec{Z}_{target})$
            for the action sequence $\cvec{u}^{k}_{0:T- 1}$ with $k=1...K$\;
            Optimize $\cvec{u}^{k}_{0:T- 1}$ with $\nabla_{\cvec{u}^{k}_{0:T- 1}} c^k$
        }
    }
    $k_{opt} = argmin_k c^k$ \;
    \Return{ $\cvec{u}^{k_{opt}}_{0:T- 1}$ }\;
\end{algorithm}
\begin{figure}[t]
\centering
\includegraphics[width=1\linewidth]{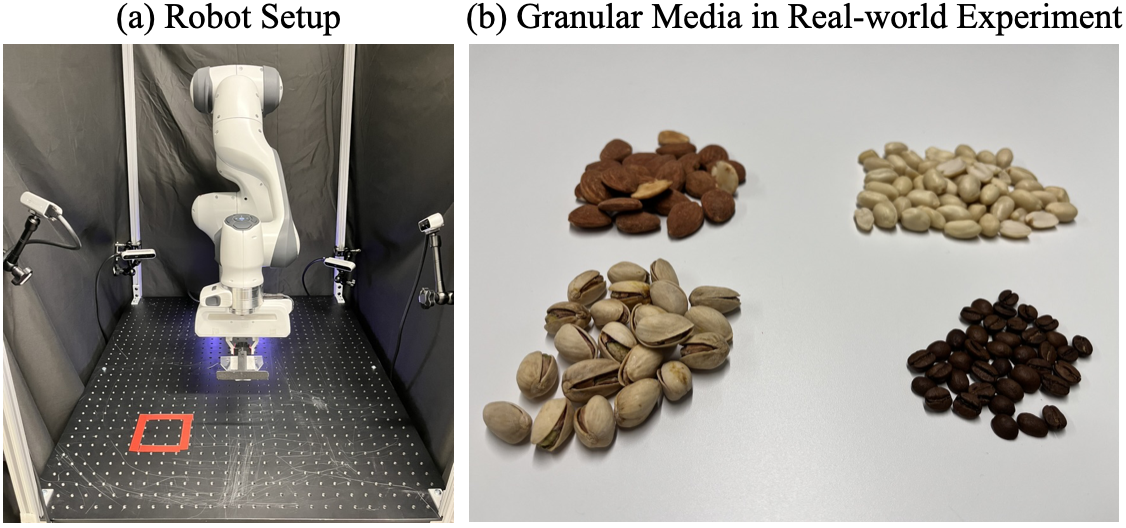}
\caption{\textbf{Real-world experiment setup}. 
\textcolor{update}{
(a) The robotic manipulator, with a pusher attached to the end-effector, moves granular materials within the workspace. Four calibrated RGBD cameras are mounted around the workspace to provide multi-view observations. 
}
(b) The granular materials used in real-world experiments include coffee beans, peanuts, pistachios, and almonds.}
\label{fig:real-world}
\end{figure}

\begin{figure*}[t]
    \centering
    \includegraphics[width=1\linewidth]{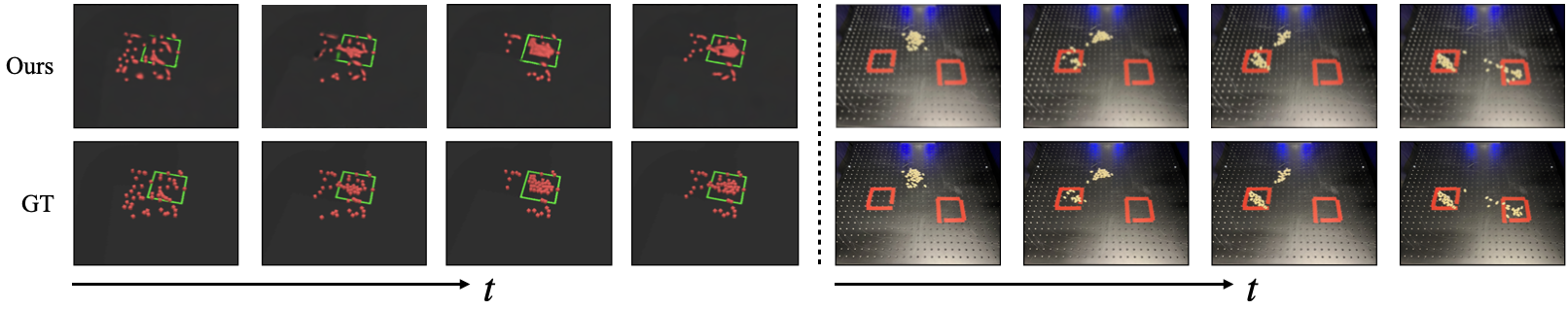}
    \caption{\textbf{Dynamics model rollouts.} We show the rollout predictions of the dynamics model in both simulation (left) and real-world data (right). Both of the rollout results show that the dynamics model prediction is accurate for a few steps.}
    \label{fig:rollout}
\end{figure*}

\section{Experimental Results}
\label{sec:result}

\subsection{Implementation Details}
We implement the entire framework using PyTorch~\cite{pytorch} and PyTorch-Geometric~\cite{pytorch-geo}.

\textbf{Simulation Setup}.
The simulation environment is based on PyBullet~\cite{pybullet}, adapted from the Ravens~\cite{corl20_transportnet} framework. Throughout both data generation and evaluation, unless otherwise specified, the granular material is formed by a set of 50 cubic blocks, each measuring 1 cm in size, along with a planar pusher of 10 cm in length.

\textbf{Physical Robot Setup}.
\textcolor{update}{Our real-world experimental setup includes four Intel RealSense cameras and a Franka Panda manipulator~\cite{arxiv20_frankapy}. More details are provided in the Appendix, found on our project website.}
The camera captures top-down views of the workspace from different angles.
We directly transfer our model trained in the simulation environment to our real-world experiment setup. 
\textcolor{update}{To alleviate the sim-to-real gap,
we resize the input images to match the workspace size used in the simulation. 
Also, we align the width of the pusher, which is attached to the robot’s gripper with its simulated counterpart
Fig. \ref{fig:real-world} shows the granular materials tested and the robot setup.
}

\textbf{Baselines}.
We compare our approach against several baselines, providing a brief description of each below:

\begin{itemize}
\item \textbf{Dyn-Res}\cite{rss23_dyna_resolution} 
\textcolor{update}{
constructs dynamic-resolution particle representations for granular materials and learns a unified dynamics model with GNN. This approach enables continuous adjustment of the abstraction level, allowing the agent to adaptively select the optimal resolution at each MPC step during testing.
}
\item \textbf{NFD}\cite{corl23_neural_field}
\textcolor{update}{
uses a fully convolutional neural network built on a density field-based representation of object piles and pushers.
}
This approach leverages the spatial locality of inter-object interactions and translation equivariance through convolutional operations.
\item \textbf{NeRF-dy}~\cite{corl21_nerfdy} 
\textcolor{update}{
leverages NeRF to learn viewpoint-invariant and 3D-aware scene representations.
%
}
By constructing a dynamics model over this learned representation space, NeRF-dy enables visuomotor control for challenging manipulation tasks, such as pouring liquid.
\item \textcolor{new}
{\textbf{DVF}~\cite{wafr20_} is a dynamics model that uses field-based state representation without inductive biases. 
}
\end{itemize}

\begin{figure*}[ht]
\centering
\includegraphics[width=1\linewidth]{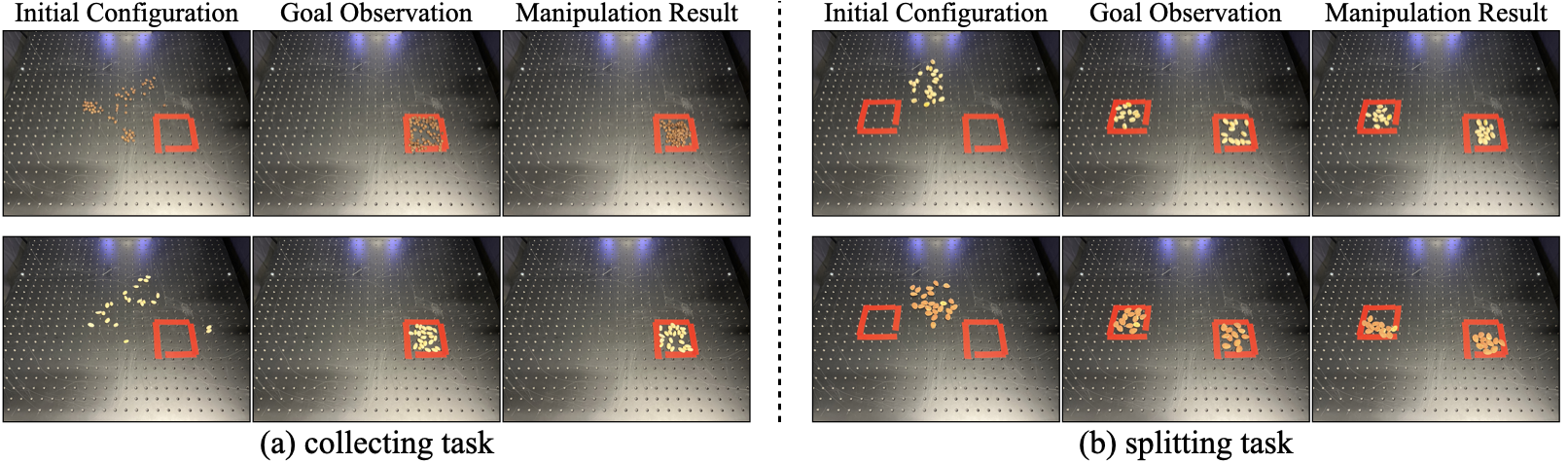}
\caption{\textbf{Qualitative results from real-world experiments.} (a) Evaluation of our method on a collection task with different objects than what it was trained on. The objects vary in scale and physical properties (e.g., almonds and pistachios remain quasi-static during MPC steps, while peanuts and coffee beans may roll after being pushed). (b) Pushing object piles into two separate target configurations. Our method successfully pushes randomly scattered objects into the desired locations.}
\label{fig:realworld_qual}
\vspace{-12px}
\end{figure*}

We evaluate the entire framework on the following tasks:
\begin{itemize}
\item \textbf{Collecting}: pushing the piles into a target region.
\item \textbf{Splitting}: pushing the piles into multiple target regions.
\item \textbf{Redistributing}: redistributing the piles to match a specific pattern.
\end{itemize}

\textbf{Metrics}.
In simulation, we perform 100 trials, while for real-world experiments, we conduct 20 trials. We use two metrics to evaluate the frameworks. 
\begin{itemize}
\item \textbf{Success rate}: success is defined as moving all materials to the target region.
\item \textbf{State error}: in simulation experiments, we also measure the Chamfer distance between the particles in the target observation and the particles after manipulation.
\end{itemize}

\subsection{Reconstruction and Dynamics Prediction Results}
%
%
%
In Fig. \ref{fig:rollout}, we show the reconstruction of the rollout trajectories with our dynamic model. We can find that our approach does capture the granular material, and the movement of the granular material is also correctly predicted by our dynamic model. We also provide more novel-view synthesis results in supplementary.

\subsection{Manipulation Results}
\begin{table}[ht]
\caption{Manipulation success rate in simulation (max = 1.0)}
\centering
    \begin{tabularx}{\linewidth}{X l l l}
    \toprule
     & Collection & Splitting & Redistributing\\
    \midrule
    \textbf{NeRF-dy}~\cite{corl21_nerfdy} & 0.67 & 0.43 & 0.31 \\ 
    \textbf{Dyn-Res}~\cite{rss23_dyna_resolution} & 0.79 & 0.72 & 0.67 \\ 
    \textbf{NFD}~\cite{corl23_neural_field} & \textbf{0.89} & 0.74 & 0.46 \\ 
    \textbf{DVF}~\cite{wafr20_} & {0.78} & 0.67 & 0.55 \\ 
    \textbf{Ours} & \textbf{0.89} & \textbf{0.79} & \textbf{0.75} \\ 
    \bottomrule
    \end{tabularx}
    \label{tab:simulation_mani}
    \vspace{-6pt}
\end{table}
\begin{table}[ht]
\caption{State error in simulation}
\centering
    \begin{tabularx}{\linewidth}{X l l l}
    \toprule
     & Collection & Splitting & Redistributing\\
    \midrule
    \textbf{NeRF-dy}~\cite{corl21_nerfdy} & 0.0096 & 0.0620 & 0.0717 \\ 
    \textbf{Dyn-Res}~\cite{rss23_dyna_resolution} & 0.0179 & 0.0533 & 0.0901 \\ 
    \textbf{NFD}~\cite{corl23_neural_field} & 0.0073 & 0.0310 & 0.0660 \\ 
    \textbf{DVF}~\cite{wafr20_} & 0.0093 & 0.0340 & 0.0919 \\ 
    \textbf{Ours} & \textbf{0.0027} & \textbf{0.0041} & \textbf{0.0081} \\ 
    \bottomrule
    \end{tabularx}
    \label{tab:simulation_state}
    \vspace{-12pt}
\end{table}
As shown in Tables \ref{tab:simulation_mani} and \ref{tab:simulation_state}, our approach consistently outperforms all baseline methods. When comparing our method to NeRF-dy, we observe significant improvements, which we attribute to our incorporation of particle clustering. This clustering allows for a better understanding of inter-particle interactions. In contrast, NeRF-dy relies on learning dynamics through NeRF reconstruction, which lacks the use of physics-based priors. NFD performs well on simpler tasks like collection and splitting but struggles with more complex target patterns.

\begin{table}[ht]
\caption{Real-world manipulation success rate (max = 1.0).}
\centering
    \begin{tabularx}{\linewidth}{X l l}
    \toprule
     &  Collection & Splitting\\
    \midrule
    \textbf{Pistachios} & 0.85 & 0.80 \\ 
    \textbf{Almonds} & 0.85 & 0.75 \\ 
    \textbf{Peanuts} & 0.85 & 0.85 \\ 
    \textbf{Coffee Beans} & 0.65 & 0.60 \\ 
    \bottomrule
    \end{tabularx}
    \label{tab:realworld}
    \vspace{-6pt}
\end{table}

In real-world experiments, as shown in Table \ref{tab:realworld}, we observe high manipulation performance across most materials. Qualitative results are presented in Fig. \ref{fig:realworld_qual}, and additional demonstrations are available on our project website.

\subsection{Generalization Studies}

In this section, we conduct ablation studies to evaluate the effectiveness of each component.

\textbf{Number of Views}.
\begin{figure}[ht]
\centering
\includegraphics[width=0.95\linewidth]{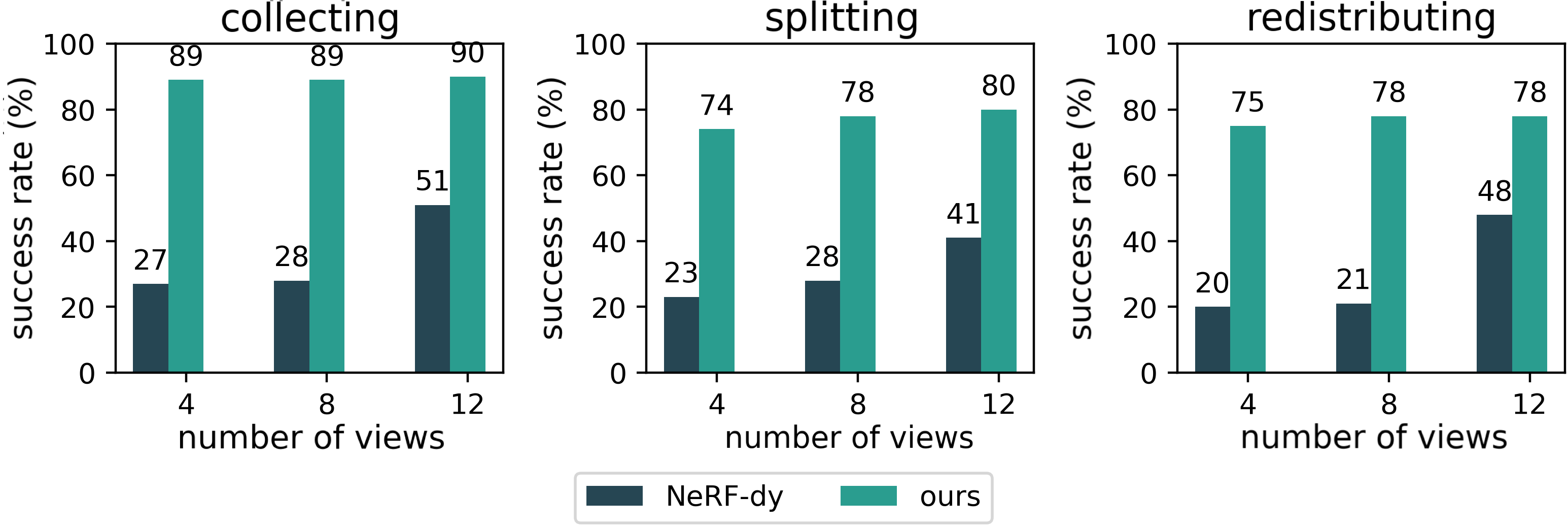}
\caption{\textbf{Manipulation performance with different numbers of viewpoints as input.} Performance increases with more views, providing more accurate granular material reconstruction.}
\label{fig:n_view}
\end{figure}
Figure \ref{fig:n_view} demonstrates that increasing the number of viewpoints improves manipulation performance by offering a more accurate reconstruction of the granular material. Our approach achieves higher performance than NeRF-dy while requiring fewer views.

\begin{figure}[ht]
\centering
\includegraphics[width=0.95\linewidth]{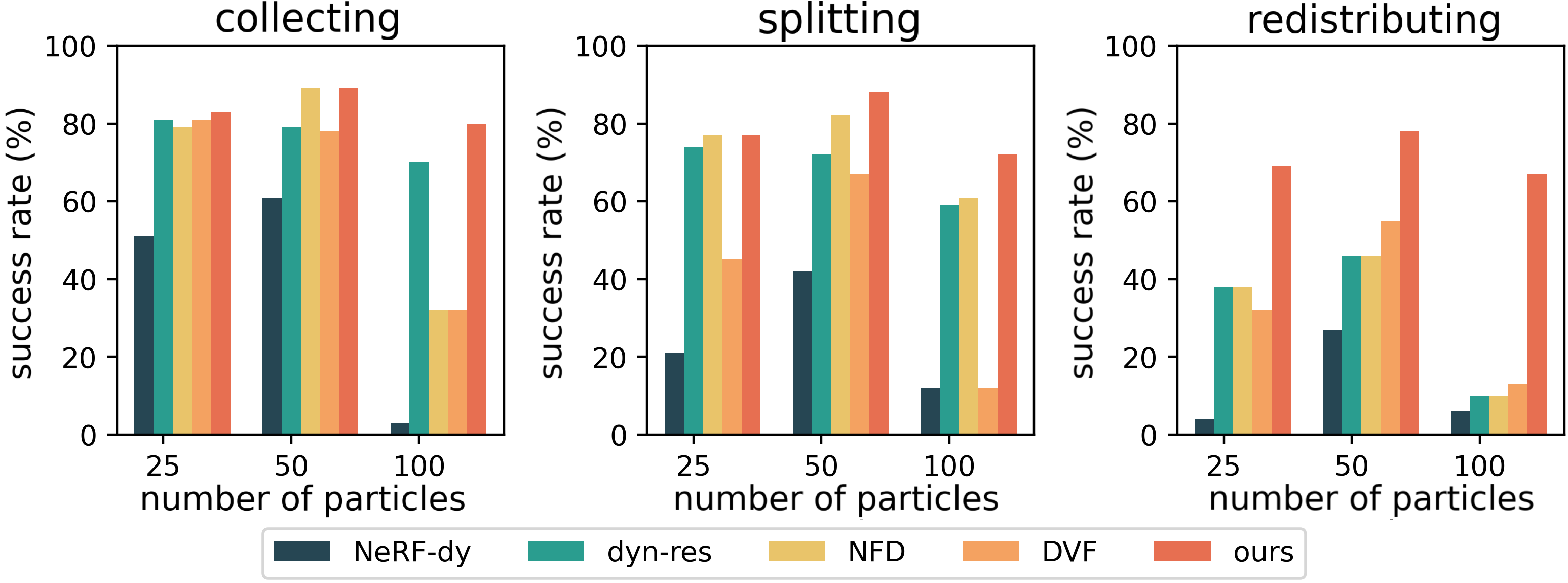}
\caption{\textbf{Manipulation performance with different numbers of particles in the workspace.} Our approach demonstrates superior generalization compared to other baselines.}
\label{fig:n_particles}
\vspace{-12pt}
\end{figure}

\textbf{Generalization to Different Numbers of Particles}.
Though our approach was trained with 50 particles, we found that it generalizes well to varying numbers of particles (Fig. \ref{fig:n_particles}), largely due to the graph neural network we use to model dynamics. In contrast, NeRF-dy and NFD, which rely on latent vectors or visual observations, struggle to generalize across different data distributions.

\textbf{Message Passing in Dynamics}.
We find that message passing plays a crucial role in capturing granular material dynamics. Tasks requiring accurate future state predictions benefit from additional message-passing steps for precise manipulation.

\begin{figure}[ht]
\centering
\includegraphics[width=0.95\linewidth]{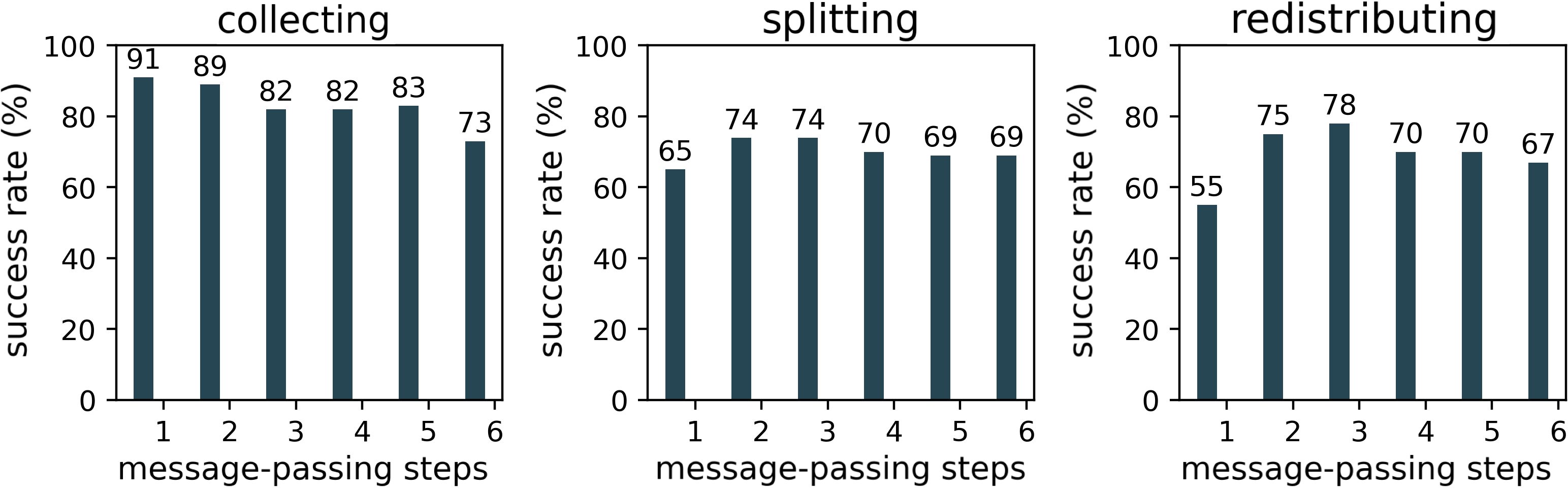}
\caption{\textbf{Manipulation performance with different numbers of message-passing steps.} More steps lead to better performance.}
\label{fig:mp}
\end{figure}

\section{Limitations}
\textbf{Manipulation Efficiency}.
Granular material dynamics are inherently complex due to the intricate interactions between particles. To manage this challenge, we limit the robot arm’s movement speed, reducing the intensity of particle interactions and promoting more stable and controllable manipulation. 

\textbf{Particle Size Limitation}. Our framework is less suited for manipulating very small particles, such as salt, sugar, or rice. This limitation stems from the difficulty in accurately reconstructing such tiny particles using Gaussian splatting, which struggles to maintain precision at smaller scales. We anticipate however that this would be an issue for any vision-based policy.

\textbf{Shortsighted Planning}. 
Despite the fact that the proposed approach effectively optimizes for the next best trajectory based on a cost function, it focuses on short-term decision-making.
%
%
In more complicated tasks, 
\textcolor{update}{
such as more particles or a complex target pile shape, 
}
planning several steps ahead may be necessary to determine truly optimal actions. 





\section{Conclusion}
\label{sec:conclusion}
This work presents a novel approach to granular material manipulation using Gaussian splatting as a latent representation. By encoding the material’s state into a probabilistic form, we effectively model and predict the dynamics of granular interactions. Our integration of this learned model with Model Predictive Control enables precise and adaptive manipulation in real-time.

Experiments demonstrate that our method significantly improves manipulation accuracy and stability over existing approaches. This highlights the potential of Gaussian splatting as a powerful tool for advancing robotic manipulation, especially in complex environments. Future work could extend this framework to other non-rigid materials, further enhancing the capabilities of robotic systems in dynamic tasks.

\newpage



\newpage



\bibliography{example}{}
\bibliographystyle{IEEEtran.bst}

\end{document}